# Can Chat GPT solve a Linguistics Exam?

Patricia Ronan, TU Dortmund University,
patricia.ronan@tu-dortmund.de
Gerold Schneider, University of Zurich,
gschneid@ifi.uzh.ch

**Abstract**
The present study asks if ChatGPT4, the version of ChatGPT which uses the language model GPT4, can successfully solve introductory linguistic exams. Previous exam questions of an Introduction to Linguistics course at a German university are used to test this. The exam questions were fed into ChatGPT4 with only minimal preprocessing. The results show that the language model is very successful in the interpretation even of complex and nested tasks. It proved surprisingly successful in the task of broad phonetic transcription, but performed less well in the analysis of morphemes and phrases. In simple cases it performs sufficiently well, but rarer cases, particularly with missing one-to-one correspondence, are currently treated with mixed results. The model is not yet able to deal with visualisations, such as the analysis or generation of syntax trees. More extensive preprocessing, which translates these tasks into text data, allow the model to also solve these tasks successfully.

Keywords: generative language model, LLMs, artificial intelligence, linguistic analysis, domain-specific knowledge, AI in studying

## 1. Introduction

The first version of the chatbot ChatGPT [1] arrived in late 2022. Since then, it has caused a quantum leap for users world-wide in the possibilities to access content produced by artificial intelligence. At the time of writing, ChatGPT4, the version of ChatGPT which uses the language model GPT4, is the state of the art program. Given its huge impact and the attention it has received, ChatGPT and related programs have an increasingly huge impact on teaching, learning and examining at educational institutions, including 3rd level institutions. The programmers of ChatGPT, OpenAI, actively support and encourage the use of the language model in educational contexts, for example by advising educators on possibilities to use the language model in education[2]. Use of the model is not only advocated for teaching, but also various internet sites point out benefits and possibilities of using AI for studying, for writing [3] and also help with assignments and essays is mentioned [4]. While such resources are very powerful and versatile tools, the surge in awareness of them naturally also leads educators not only to adapt their own teaching towards the inclusion of such tools, but also to worries about in how far AI applications might be submitted instead of original student work. Coming from the discipline of linguistics, the authors of this study were thus interested in finding out whether ChatGPT would be able to solve a basic linguistics exam at the Technical University of Dortmund in Germany. In order to investigate this, we gave the questions of the course exams of the first part of an Introduction to Linguistics in the winter 2017/2018 to ChatGPT, in order to find out whether ChatGPT could solve the exam. At the time, the graded coursework consisted of two take-home assignments and one in-class test. The tests pertaining to the first component of the Introduction to Linguistics, Linguistics 1 test knowledge on the linguistic fields of Morphology, Phonetics and Phonology and Syntax. Further subfields of linguistics, namely semantics, pragmatics, sociolinguistics, historical linguistics or language acquisition are the topic of the Linguistics 2 course at the Technical University of Dortmund, and have not been dealt with in this study.

In the following, we will briefly introduce relevant background research on ChatGPT, before we outline our data and method and describe the results of the study.

## 2. Background

ChatGPT is a variant of a Generative Pretrained Transformer model, or GPT. ChatGPT is a specific application of the language model that has been designed for chatbots. The models use transformer blocks, which comprise neural networks, to learn relationships between words in sentences in order to (re-)create naturalistic content [5]. The model is trained by self-supervised learning on very large amounts of text, including entire webdumps, the entire Wikipedia etc., also known as large language models (LLMs). While the one task that these models learn, namely to guess missing words like in a cloze test, at the first seems remote from actual tasks to be solved, they have led to a quantum leap in AI. First, the very large amount of data injects generic world knowledge into the models. Second, specific training, so-called fine-tuning, with supervised learning can also be added, for instance for question-answering or translation [6]. As such, the model can also be used in various academic fields [7] and particularly much research on the use of ChatGPT exists in the fields of Computation and Language, but also in Artificial Intelligence, Machine Learning and Computers and Society. Yiheng Liu et al. [7] note that it also shows "potential" in wider research fields, particularly in education, but also in physics and medicine. It can further be used for text classification or text or code generation and for drawing inferences. While Jagdishbhai and Thakkar [6] caution that ChatGPT is currently restricted to textual data and cannot yet deal with other data types such as audio, video or images, Yiheng Liu et al. [7] show that research on data visualization and multimodal data processing with ChatGPT is starting.

Jagdishbhai and Thakkar [6] further point out that the use of ChatGPT has various advantages and disadvantages. They stress that ChatGPT is constantly available and immediately accessible, that it is fast and efficient, that, as an artificial intelligence model, it does not have human biases, and that it can communicate in various languages, which allows for wide international accessibility. Disadvantages, by contrast, are that the knowledge of the model is restricted to what it has been trained on. It further lacks human intelligence and empathy, even though it has also been pointed out [8] that, when the Turing test is applied to it, the model itself considers itself sentient, and actually passes the Turing test. Further disadvantages of ChatGPT are that it does not necessarily know the full context of a question, it may give wrong responses and finally, if it has been trained with unreliable or outdated data, it may also give incorrect answers. In source texts with long-range dependencies, especially in scientific or legal texts, data processing may be difficult [6], Yiheng Liu et al. [7] further caution that next to the danger of using outdated knowledge and lack of understanding, such large-scale models also have high energy consumption and carbon emissions, and they can be used maliciously when users overcome blocks avoiding model toxicity. Further, political and ideological biases can be observed, and there are concerns about data security in connection with the language model.

Currently evaluation of the language models is typically by measures of perplexity or by evaluation of the generated texts [6]. This second method is adopted in the present contribution, which evaluates the output created by ChatGPT 4 on the questions of an Introduction to Linguistics final course exam at a German university.

## 3. Data and Method

This study investigates the performance of ChatGPT4 on the exam questions of the final exams of an Introduction to Linguistics course. The final exam of the first semester of the Introduction to Linguistics contains questions on Syntax, Morphology, Phonetics and Phonology, and the questions from the exam session of the winter 2017/2018 exam were fed into ChatGPT, Version 4. For doing so, only minimal preprocessing of exam questions was needed. Two additional steps were implemented. First, strings containing the verbatim examples on which questions were based were marked by [X is] and [In X], e.g.

- X is "This is an example"
- In X, determine all parts of speech

Second, where there was underlining in the original exam questions, this was replaced by square brackets. The output of ChatGPT was then corrected by two examiners according to the normal class grading scheme.

## 4. Findings

As noted in 3. above, the exam with which ChatGPT was tasked consisted of different parts: a morphology section, a syntax section and a phonetics/phonology section. ChatGPT performed differently in these different parts. In the following, we will discuss its performance.

### 4.1 Morphology component of the exam

The morphological component of the exam comprises different tasks including determination of parts-of-speech, of the analysis of morphemes and the identification of lexemes. The determination of parts-of-speech (Figure 1) and the identification of lexemes were fairly successful, the morphological analysis (Figure 2) was less so.

In the statement you provided (X), let's identify the lexical categories (parts of speech) of the underlined words:
"Jeremy Hunt has said giving MPs a vote on Theresa May's Brexit negotiating plans would harm the economy, but parliament was 'highly likely' to get a vote on the final deal."

- Jeremy Hunt: Proper noun (personal name) v
- has said: Verb phrase (verb + auxiliary verb) *has* is the auxiliary – this is not clear here
- giving: Verb (gerund form of the verb "give")
- MPs: Noun (acronym for "Members of Parliament") v
- a vote: Noun phrase (article + noun)
- on: Preposition
- Theresa May's: Adjective (possessive form of the proper noun "Theresa May")
- Brexit negotiating plans: Noun phrase (adjective + noun + noun)
- would harm: Verb phrase (auxiliary verb + verb) v
- the economy: Noun phrase (article + noun)

Figure 1: Performance of ChatGPT in the analysis of parts-of-speech with annotators' comments in red colour. 'v' is short for the tick symbol indicating correct answer.

The determination of parts-of-speech was successful overall. However, difficulties could be observed with the classification of adjectives versus adverbs and with the determination of auxiliaries. In the phrase *has said*, the distribution of auxiliary and main verb is not clear and possibly incorrect. Further, in the adjectival phrase *highly likely,* the adverb *highly* and the adjective *likely* are classified as adjectives. In answer to the question whether any lexemes are ambiguous in their classification, the gerund *giving* is correctly identified as a potentially ambiguous category.

In a further task, the word-formation processes in the short text given in Figure 1 were to be identified. This task is largely carried out correctly by ChatGPT, and it correctly identifies blending (*Brexit*), suffixation (*negotiating, highly*). Again, the analysis of the phrase *is highly likely* is unsuccessful, however. Applying somewhat faulty logic, ChatGPT determines that *likely* ‘is formed by adding the suffix "-*ly*" to the adjective "*likely*."’ This analysis arguably shows that the model, as argued by Yiheng Liu et al. (2023) indeed lacks the cognition of a (linguistically trained) human cognizer. The solution of the following explanation task, the definition of the term backformation, by contrast, is a core-competence of ChatGPT and is carried out very well.

Predictably, much more mixed results are found if ChatGPT is asked to make a morphemic analysis of different words into free or bound and lexical or grammatical morphemes. Certain morphological core-categories are analysed quite successfully. This holds especially for instances of the plural morpheme *–s.* Thus *hats* is analysed correctly, whereas the model does not have sufficient insights for slightly more complex morphological categories. Figure 2 shows that in the token *disagreeably,* the prefix *dis-* is misanalysed as a grammatical morpheme, rather than as a lexical morpheme, as it should have been. The model is least successful in its analysis of *disdains* and *Tyrolese.* Having identified *dis-* as a potentially detachable prefix, *disdain* is misanalysed as consisting of a prefix and a stem, which is not the case as *dain-* cannot be a possible verb-stem on its own. The high-frequency third person singular present tense –*s* ending, by contrast, is correctly analysed by the model.

Let's isolate the morphemes of the words given and determine their types and statuses in the word-formation process:

1. disdains:
   - dis- (prefix) [bound, grammatical] v, not detachable from –dain here, also not grammatical but lexical.
   - dain (root) [free, lexical] no, not in contemporary English v
   - -s (suffix) [bound, grammatical] v, v, v
   - stem: disdain (root + suffix) v

3. Tyrolese:
   - Tyro- (root) [free, lexical] no, tyrol, v
   - -lese (suffix) [bound, lexical] v, v, v
   - stem: Tyro (root) no

Figure 2: ChatGPT’s performance in the analysis of word-formation

The difficulties with the analysis of these lexemes confirms the observation [6] that ChaptGPT performs badly on rare and unusual words, and that its performance may be constrained by a lack of comprehensive world knowledge and missing human cognizer competence of ChatGPT [7].

Lack of world-knowledge, especially of rare words or concepts, also leads to restrictions in the last morphology exercise, the determination of the number of different lexemes in the following tongue twister: *A bitter biting bittern Bit a better brother bittern, And the bitter better bittern Bit the bitter biter back. And the bitter bittern, bitten, By the better bitten bittern, whined and viciously bit him in his back.* Here, the two different lexemes of *back* are not distinguished. Further, the lexeme *biter,* an agent-noun, is equated with the lexeme *bit,* a verb in the past tense. Surprisingly, the indefinite article *A*, spelled in capital at the beginning of the clause, is not correctly recognized as a spelling variant of the indefinite article *a* later in the same clause.

Thus, overall, ChatGPT delivers a, in some respects, reasonably good morphological analysis, especially where it can create essay-type answers for questions on morphological issues, but it does not yet have the ability to carry out morphological analyses which go far beyond the analyses of inflections and of common and frequent derivational patterns. The two examiners of the exam would have assigned an overall 19.5 of 25 possible points to this part of the exam, ChatGPT would thus have passed the morphology part of the test.

**4.2 Phonetics and Phonology component of the exam**

The Phonetics and Phonology part of the test is solved remarkably successfully by the language model. Again, not surprisingly, the answers to the knowledge-based questions of the exam are extremely well solved, the language model has received full points for its answers to each of the questions by both annotators:

1. Name and describe the three different types of phonetics
2. Name four factors that determine vowel quality
3. Provide an articulatory phonetic description of the sounds represented by the symbols [ə], [n], [æ], [t], [ʊ], [tʃ], [f]
4. Transcription of items in broad phonemic transcription into normal English spelling

As can be seen in Figure 3, ChatGPT was able to draw on enough knowledge to correctly carry out the transformation of broad transcription of IPA (i.e. International Phonetic Alphabet) signs into English words. It also manages to largely cope with the internal exercise structure of exercises a) to j). However, in its output, an extra tabulator from the manuscript of the exams in front of answer j) confuses the index in the output of the data.

Please transcribe the following items in broad phonemic transcription into normal English spelling

a) /tʃi:t/  f) /əˌsɪmɪleɪʃən/
b) /sʌn/  g) /lɪŋ'gwɪstɪks ɪz fʌn/
c) /baɪt/  h) /fə'netɪks tu:/
d) /θɔ:t/  i) /raʊn ðə kɔ:nə/
e) /əʊld ‘ɪŋglɪʃ/  j) /ɪ'nʌf /

Here are the transcriptions of the items in broad phonemic transcription into normal English spelling:
a) cheat
b) sun
c) bite
d) thought
e) old English
f) assimilation
g) linguistics is fun
h) phonetics too
i) round the corner
e) enough
Please note that the transcriptions provided are based on a broad phonemic transcription, which represents the phonemic contrasts in a language rather than the precise phonetic details. As a result, the transcriptions may not capture regional variations or specific phonetic nuances.

Figure 3: Phonemic transcription task and ChatGPT solutions

ChatGPT was similarly successful in transcribing a short English language text into broad phonemic transcription including weak forms. The only shortcoming was that in some instances weak forms were not correctly rendered as weak forms but as fully stressed forms. Overall, the two graders have assigned 24.5 points of the possible total of 25 points to ChatGPT’s answers to the phonetics and phonology part of the linguistics exam. Here, it seems particularly remarkable that in addition to the definitions that may have been obtained from data on which the language model was trained, ChatGPT has also shown itself able to correctly transcribe written English text into IPA transcription and vice versa. In this task it has performed better than in the morphological analysis: evidently the model was able to draw on and employ the analytic toolkit for phonetic transcription more successfully than the tools for morphological analysis. This is likely to be due to the fact that there is a more explicit and more restricted one-to-one correspondence between item and category in phonetic transcription than in morphological analysis.

### 4.3 Syntax component of the exam

Trends that have been visible in Sections 4.1 and 4.2 above continue to be observable in the answers to the questions in the syntax part of the exam which have been provided by ChatGPT. Again, the terminology which had to be explained was defined correctly. In the analytical part of the syntax questions, the model performed less well. The first task in the exam was to identify syntactic phrases. The task was to decide whether a word in a [ strings of words ] in angular brackets constitute a phrase or not. Further, the result was to be supported, i.e. motivated, in each case by applying two phrase identification tests. Given the lack of one-to-one correspondence between words and possible phrases, it may not be surprising that ChatGPT had some difficulties in identifying the phrases of the given text correctly. For this, five sentences were presented and it had to be determined whether bracketed elements constituted a syntactic phrase or not. The sentences in question were 1. *Clarice [gave a doughnut] to Samantha yesterday,* 2. *This can wait until [the very last day],* 3. *She went to the children [with chocolate*], 4. *She went [to the children with chocolate]*, and 5. *[Taking a bath] always made Venus feel better.*

ChatGPT analyses all items in angular brackets as correct syntactic phrases. In each case, this is motivated in the answers by claiming that either a replacement test of phrasal status can used, for example, that the item in question can either be replaced for example by a pronoun (e.g. in sentence 2 *it* can replace [*the very last day*], in sentence 5, [*Taking a bath*] can be replaced by *it*. Further, ChatGPT claims that subject or object status of the "phrase" determines is phrasal status. The former, the replacement test, is a bona-fide test of phrasal status, the latter, subject- or object status, is not, and neither is the assignment of subject- or object status to presumed phrases is correct in all cases (see Figure 4).

1. Clarice [ gave a doughnut ] to Samantha yesterday. Phrase identification tests: a. Can it function as a single unit in a sentence? - YES (It can be the subject or object of a sentence.) b. Can it be replaced by a pronoun? - YES ("She" can replace "Clarice gave a doughnut to Samantha yesterday.") Result: YES, it is a phrase. The phrase formed is "gave a doughnut."

No, and the phrase is not correctly understood.

Figure 4: ChatGPT answers to a phrase determination task and graders' comment.

Figure 4 illustrates that ChatGPT misanalyses *give a doughnut* as a syntactic phrase, and considers it to be an object. The correct use of movement tests, the attempt to ask a question for this element or to try and make the element the predicate of the *it* – clause in a cleft sentence would have obtained better and correct results (cf. e.g. [9]). The suggestion that *she* can replace the whole sentence is obviously wrong, and a type of mistake that human cognizers would hardly make.

Equally mixed were the results for an analytic task in which specific linguistic items had to be found in a short text that was provided in the exam. While some more frequent features were correctly identified, other concepts were clearly too complex for a successful analysis (Figure 5).

ChatGPT

Linguistics description Example

1. a complex-transitive verb - "had emphasized" v
2. a phrase that consists only of its head - "Private inquiry" f
3. a clause that functions as an NP modifier - "that they disliked nosy questions" f
4. an adverbial - "always" v
5. a direct object that is not realized by an NP - "nosy questions" f
6. an instance of a ditransitive construction - "she might have forgiven Annie her violent reaction" v
7. an indirect object - "Annie" (in "she might have forgiven Annie her violent reaction") v
8. a subject complement - "more emotional" (in "would make their work situation a lot more emotional") f

2 of 4 points

Figure 5: ChatGPT output identifying linguistic items purportedly matching linguistic descriptions with graders' comments in red.

The selection of items from a text fitting the given linguistic description has some correct points, especially with items of high or fairly high frequency, such as an adverbial (*always*), an indirect object (*Annie*), or, more difficult, a complex transitive verb (*had emphasized*) or a ditransitive construction (*forgiven Annie her violent reaction*). However, questions asking for more complex concepts like heads or complements, or a question that excludes a feature (the negated line 5) lead to an error here. Thus, the tasks to find a phrase only consisting of its head fails (*private inquiry* clearly has a modifier), the object that is not supposed to be realized by an NP is clearly realized by one (*nosy questions*). Possibly, ChatGPT lacks the domain-specific terminological knowledge of what a linguistic head is, a concept that is far remote from everyday uses of the word *head*. Generally, in light of the results in Sections 4.1 and 4.2, this outcome is not surprising as this analysis of unseen content could not have formed part of a training database and needed transfer skills for the analysis of more complex questions. In this respect, this task resembled the morphological analysis illustrated in Figure 2 and while it is remarkable that ChatGPT can apply some analytic steps correctly, in others human cognition is missing.

The next task to be carried out was the analysis of a syntax tree. This task could not be undertaken by ChatGPT 4 at all without major pre-processing because the model without any extensions was not yet able to analyse the visual component of the syntax tree (cf. [6]). As the task thus was not fulfilled, the graders decided to give 0 points for this task. Alternatively, when asking the question, extensive pre-processing by changing the visual data into written text format could have been implemented to make the question easier for ChatGPT to handle. This, however, would have been part of the task of the following linguistic analyses and the graders decided not to award any points.

Similarly, an exercise to draw a syntax tree of a given sentence fails for the same reason: the model cannot (yet) engage with multimodal data production, here the drawing of syntax trees. In this case, by contrast, the language model engages with the task by providing a syntactic analysis that does not comprise the asked-for syntax tree (Figure 6).

5.1 Draw a syntax tree for the following sentence: Rey returned the lightsaber to Luke Skywalker in an unmapped part of the galaxy.

ChatGPT

As an AI text-based model, I can't literally draw diagrams. However, I can describe the structure of the syntax tree for the sentence you've provided.

Here is a step-by-step breakdown of how you might construct a syntax tree for the sentence "Rey returned the lightsaber to Luke Skywalker in an unmapped part of the galaxy."

1. The whole sentence is the root of the tree and is labeled as a Sentence (S).
2. Divide the sentence into its main parts: The subject "Rey", the verb phrase "returned the lightsaber to Luke Skywalker in an unmapped part of the galaxy." This gives us two branches from the root:
    - Noun Phrase (NP) for "Rey",
    - Verb Phrase (VP) for "returned the lightsaber to Luke Skywalker in an unmapped part of the galaxy."

Figure 6: ChatGPT providing hierarchical textual output instead of sentence tree.

Provided the graders are willing to accept a non-tree output for this task, there are some correct items in this analysis. The subject noun phrase *Rey* and the verb phrase *returned the lightsabre to Luke Skywalker in an unmapped part of the galaxy* are correctly identified. Further nodes in the sentence are also correctly identified. Thus, if the grader is happy to accept non-tree solution, ChatGPT's output on this exercise is acceptable overall. A further subsection of the question, "How would the tree look for the phrase *The lightsabre that Rey returned*", was subject to the same constraints (Figure 7). Again, the analysis of visualisations still poses severe problems for the language model [6], which severely impacts its ability to deal with syntax trees actively or passively.

How would the tree look for the phrase "The lightsaber that Rey returned."

This sentence structure can be analyzed as follows:

The sentence ("S") splits into a Noun Phrase (NP, "The lightsaber that Rey returned").

The NP can be further broken down into a Determiner (D, "The"), a Noun (N, "lightsaber"), and a Relative Clause (RC, "that Rey returned").

The RC splits into a Relative Pronoun (RP, "that"), a Noun (N, "Rey"), and a Verb (V, "returned").

So, if you were to draw a syntax tree based on this breakdown, it would look something like this:

- S
  - NP
    - D: "The"
    - N: "lightsaber"
    - RC
      - RP: "that"
      - N: "Rey"
      - V: "returned"

Again, keep in mind this is a text-based representation and actual syntax trees are visual diagrams. You could use an online syntax tree generator to create a visual version of this tree.

Figure 7: ChatGPT analysis of *The lightsabre that Rey returned*

Overall, for the syntax part of the exam, definition tasks are again well-solved, the phrasal analysis, however, is only partly successful and shows partly haphazard results beyond simple and frequent categories. But so far, the biggest problem for the language model is the missing possibility to engage with syntax trees both actively and passively: neither can they be analysed nor produced. Overall, ChatGPT manages to obtain 13.75 points for the syntax part of the exam, which is below pass level for the syntax component itself.

In summary, taking the results for all the constituent parts of the exam together, ChatGPT managed a total of about 77% correct results in the Linguistics 1 exam. Of the possible total of 75 points, the graders awarded 57.5 points, with ChatGPT performing best in the phonetics and phonology part of the exam (24.5 of 25 possible points, or 98%), and worst in the syntax part of the exam (13.75 of 25 possible points, 55%), and close to average in the morphology part (19 of the 25 possible points, 76%). With the overall result, ChatGPT performed slightly above the class average, which was at 72% (standard deviation of 9.5). ChatGPT thus reached an overall fairly good pass level of a first-year Linguistics 1 test.

## 5. Conclusion

The question whether ChatGPT can pass a linguistics exam can receive a clear *yes* as an answer. Overall, the language model manages very well to interpret and correctly answer even complex and nested tasks. In particular, it performs very well in the area of phonetics and phonology, and it provides very good phonetic transcriptions. By contrast, it cannot (yet) analyze morphemes or phrases reliably. Tasks that involve the analysis of linguistic material and assigning data labels to complex entities or to entities without a one-to-one correspondence is taxing for the model. It performs sufficiently well in simple cases, but not in the more complex cases, particularly those provided by morphological or syntactic analysis. As yet, any tasks involving visualisations cannot be dealt with: tasks involving syntax trees or the interpretation of any visual data cannot be carried out by the model, while its core competency obviously is the creation of essay-like text. But, as this study has shown, the model also performs sufficiently well in the analysis of linguistic data from the point of view of different linguistic categories. Once multimodal approaches are implemented, the model is likely to be able to also deal with the analysis – and production – of visualization, such as syntax trees. ChatGPT also seems to perform less well on tasks requiring highly domain-specific terminology, such as for syntactic theory. It also performs less well on rarer phenomena due to sparse data problems. The implementation of more detailed knowledge allowing the model to deal more successfully with syntactic or morphological analysis seems possible and likely in the near future, as its good performance in the one-to-one or one-to-few correspondence tasks in phonetic transcription shows.

Overall, ChatGPT reaches human-level competence and performance without any specific training for the task and has performed similarly to the student cohort of that year on a first-year linguistics exam. Further fields of linguistics, semantics, pragmatics, sociolinguistics, historical linguistics or language acquisition have not been dealt with in this study as they are the topic of the Linguistics 2 course and not of the Linguistics 1 course that has formed the basis for this study. So far, a similar outcome for a corresponding exam task on these fields of linguistics seems likely.